\documentclass[journal]{IEEEtran}
\makeatletter
\def\endthebibliography{%
	\def\@noitemerr{\@latex@warning{Empty `thebibliography' environment}}%
	\endlist
}
\makeatother

\usepackage{cite}

 \usepackage[pdftex]{graphicx}

\ifCLASSINFOpdf
\else
\fi

\usepackage{algorithmic}
\usepackage{algorithm}

\usepackage{array}
\newcolumntype{K}[1]{>{\raggedright\arraybackslash}m{#1}} 
\newcolumntype{X}[1]{>{\centering\arraybackslash}m{#1}} 
\newcolumntype{E}[1]{>{\raggedleft\arraybackslash}m{#1}} 

\begin{document}
%
\title{Rapid Flow Behavior Modeling of Thermal Interface Materials Using Deep Neural Networks}
%
%
%

\author{Simon~Baeuerle, Marius~Gebhardt, Jonas~Barth, Andreas~Steimer and Ralf~Mikut
\thanks{The work of R. Mikut was supported by the Helmholtz Association’s Initiative and Networking Fund through Helmholtz AI. \textit{(A. Steimer and R. Mikut contributed equally to this work.) (Corresponding author: S. Baeuerle.)}}
\thanks{S. Baeuerle and R. Mikut are with the Institute for Automation and Applied Informatics (IAI), Karlsruhe	Institute of Technology (KIT), D-76344 Eggenstein-Leopoldshafen, Germany. (e-mail: simon.baeuerle@kit.edu)}
\thanks{S. Baeuerle, J. Barth and M. Gebhardt are with the Robert Bosch GmbH, D-72762 Reutlingen, Germany.}
\thanks{A. Steimer is with the Bosch Center for Artificial Intelligence (BCAI), Robert Bosch GmbH, D-71272 Renningen, Germany.}
}

%
%

\markboth{}%
{Shell \MakeLowercase{\textit{et al.}}: Bare Demo of IEEEtran.cls for IEEE Journals}
%



\maketitle

\begin{abstract}
Thermal Interface Materials (TIMs) are widely used in electronic packaging.
Increasing power density and limited assembly space pose high demands on thermal management.
Large cooling surfaces need to be covered efficiently.
When joining the heatsink, previously dispensed TIM spreads over the cooling surface.
Recommendations on the dispensing pattern exist only for simple surface geometries such as rectangles.
For more complex geometries, Computational Fluid Dynamics (CFD) simulations are used in combination with manual experiments.
While CFD simulations offer a high accuracy, they involve simulation experts and are rather expensive to set up.
We propose a lightweight heuristic to model the spreading behavior of TIM.
We further speed up the calculation by training an Artificial Neural Network (ANN) on data from this model.
This offers rapid computation times and further supplies gradient information.
This ANN can not only be used to aid manual pattern design of TIM, but also enables an automated pattern optimization.
We compare this approach against the state-of-the-art and use real product samples for validation.
\end{abstract}


%
\IEEEpeerreviewmaketitle

\section{Introduction}
\begin{figure}[!t]
	\centering
	\includegraphics[width=.9\columnwidth]{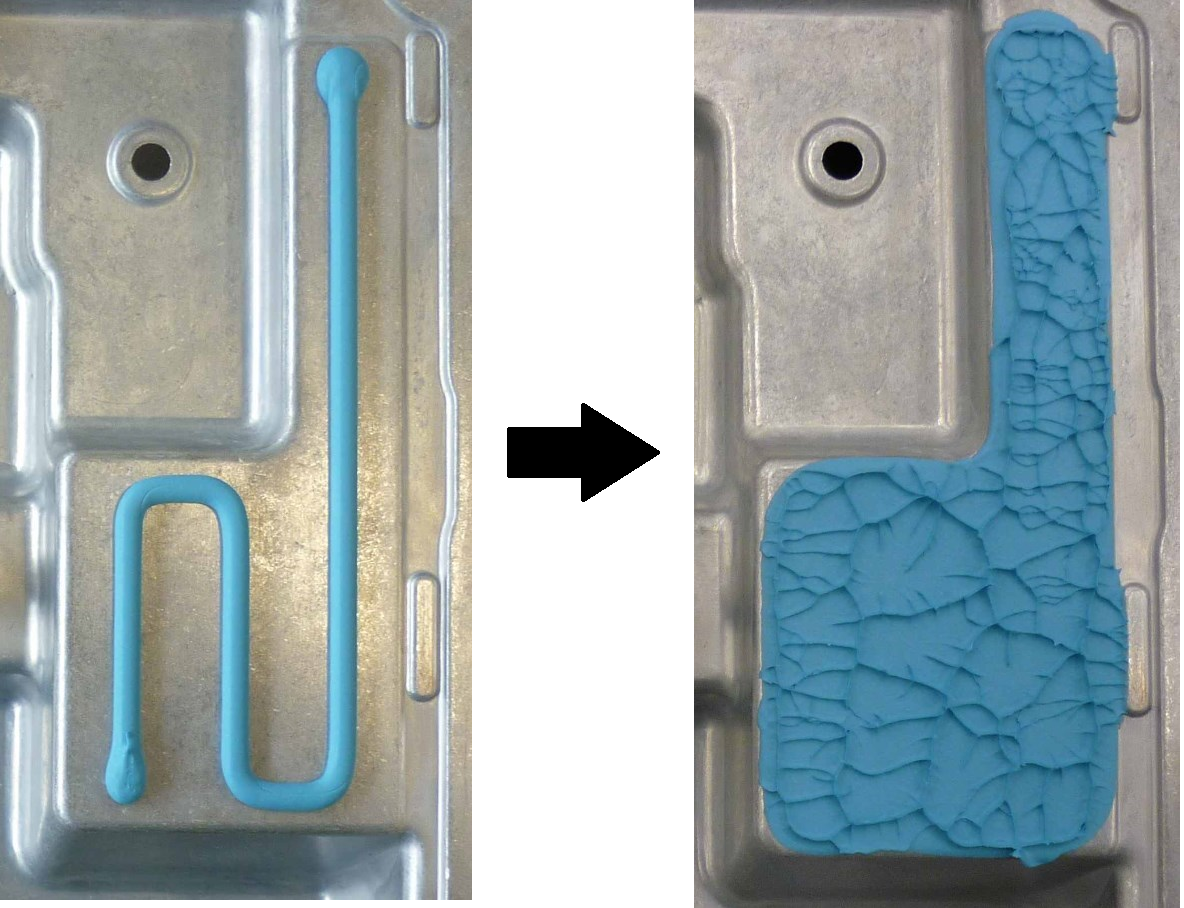}
	\caption{Material flow of Thermal Interface Material (TIM) during joining the heatsink of an Electronic Control Unit (ECU). Left: state before joining, right: state after joining.}
	\label{fig:overview}
\end{figure}

%
%
%
%
\IEEEPARstart{A}{utomotive} industry is putting an increasing effort into electric and autonomous vehicles.
Demand for efficient and reliable electronic components is rising accordingly.
This is valid for small Electronic Control Units (ECUs) used to control e.g. the engine, but also for power electronics components such as inverters or chargers.
Time-to-market tends to be shorter and is a crucial factor for global competitiveness.
While power ratings increase, tight restrictions are imposed on assembly space as well.
Thus, the thermal performance is a crucial factor in electronic packaging. \\
Thermal Interface Materials (TIMs) are widely utilized to lower the thermal resistance between individual components and thus enable an efficient heat transfer.
Typically, they are applied onto a cooling surface with a dispensing machine.
An example of dispensed TIM is shown on the left-hand side of Fig.~\ref{fig:overview}.
During the joining process of the heatsink, TIM is compressed and spreads over the surface.
The state after compression is shown on the right-hand side of Fig.~\ref{fig:overview}.
Design engineers determine the pattern, along which TIM is dispensed.
They need to consider multiple aspects.
The most evident is a \textit{high area coverage ratio} of the cooling surface with TIM after the joining process.
Along with the high heat conductivity of TIM, this results in the aforementioned low thermal resistance.
However, applying too large amounts of TIM, with excess material flowing beyond the cooling surface, leads to preventable \textit{material cost}.
This is especially relevant in high-volume series production, where even a small amount per part adds up to significant costs.
Sensitive electronic components or product features such as screw holes may be placed close to the cooling area.
They are regarded as \textit{taboo zones} during the design process and may not be covered by material. \\
A low coverage may be caused by simply applying too little TIM.
Another cause for low coverage are air entrapments, which develop during the joining process.
Air within closed contours such as circular shapes cannot escape while TIM is being compressed.
The formation of such \textit{voids} depends on the individual pattern shape and may not be easily recognized in all cases.
Furthermore, the design of a dispense pattern is directly linked to the \textit{cycle time} of the respective dispense process. \\
Apart from the pattern, the design process itself needs to be optimized as well.
The design process is relevant for the time-to-market and thus needs to be short.
The cost of human experts is also a relevant factor.
Recommendations regarding optimal patterns can help engineers during their work.
Specific guidelines exist for simple cooling area geometries such as rectangles~\cite{licari_adhesives_2011}.
However especially for larger and more complex surfaces, the dispense pattern needs to be adjusted for each individual product. \\
To evaluate a given pattern, the respective compressed state after joining needs to be known.
It can be acquired by simulating the flow behavior of TIM.
Computational Fluid Dynamics (CFD) simulations, carried out by highly specialized experts, are widely used.
Modeling and evaluation take time, but yield very accurate results.
Besides simulations, mechanical experiments are carried out with real product samples.
In an iterative fashion of trial-and-error, the dispense patterns are optimized by development engineers.
After several trials, a dispense pattern is defined.
However, mechanical tolerances are prevalent in real products.
When joining parts together, tolerances from multiple parts add up.
The high accuracy of CFD simulations, which is achieved with high efforts, needs to be weighed against the variations of real products.
A light-weight model with a lower accuracy but faster setup and computation times can fulfill the demands of dispense pattern design better. \\
CFD simulations of this kind are computationally expensive.
Artificial Neural Networks (ANNs) can be trained on data from simulation models.
They can typically be executed much faster.
Since a rather high number of training samples is needed, an automated simulation setup is typically necessary. \\
Furthermore, ANNs or other Machine Learning models are well suited to build Digital Twins, since they are generally fit to be fine-tuned by using real-world data.
They can support many design and production processes, e.g. for pose estimation from image data~\cite{baeuerle_cad--real:_2021} or for quality monitoring in resistance welding~\cite{zhou_machine_2022}.
Digital Twins are considered to be a \textit{significant enabler for Industry 4.0 initiatives}~\cite{chiabert_digital_2018}. \\
In this work, we present two flow behavior models for TIM.
They can be used to support development engineers both during manual dispense pattern design and by enabling an automated dispense pattern optimization.
The first flow behavior model is a light-weight heuristic.
The second one is an ANN.
The heuristic can be used to analyze a large range of different dispense patterns automatically.
Thus, training data can be generated for the ANN, which offers an even higher computational speed.

The remainder of this paper is organized as follows.
Section~\ref{sec:relatedwork} provides an overview over the state-of-the-art methods, which are relevant for the dispense pattern design process.
In Section~\ref{sec:simulation_methods}, we give a detailed insight into our light-weight heuristic.
It further includes the extension with an ANN and specific details on spatial resolution and the training setup.
The experiments, which we carry out to validate both of our models, are described in Section~\ref{sec:experimental_setup}.
Results in Section~\ref{sec:results} include a study of the achieved computation speed and the accuracy both on samples from the laboratory and on a real product.
Advantages and limitations of the heuristic itself as well as the combination with an ANN are discussed in Section~\ref{sec:discussion}.
%
 
\hfill August 8th, 2022
\section{Related Work}\label{sec:relatedwork}
Several works have highlighted that a high surface coverage with TIM enhances thermal performance of an electronic package.
Ekpu et al.~\cite{ekpu_effects_2012} set up a numeric simulation model including a chip, a heatsink and a TIM layer between both.
They analyze the influence of TIM area coverage on thermal resistance.
They report a lower thermal resistance with higher coverage percentages and recommend a coverage ratio of at least 75\,\%. 
They anticipate that their results will aid design engineers.
Kesarkar et al.~\cite{kesarkar_how_2019} also set up a numeric simulation model.
Their model reproduces the thermal management problem as found in an ECU, with a TIM layer below a heat sink.
They analyze different TIM coverage percentages in various configurations and report a better thermal performance for a higher TIM area coverage.
Gowda et al.~\cite{gowda_voids_2004} state that \textit{the negative effect of [...] voids on the thermal resistance of a TIM layer can be devastating}. \\
CFD simulation is a powerful tool to support design engineers during the design of dispense patterns.
They have been used in the past both to model thermal performance and to model the flow of fluid materials.
For example, Lee et al.~\cite{lee_investigation_2000} analyze both the heat conductance within a thermal package and the heat transfer to ambient air and compare different techniques to enhance heat dissipation.
Comminal et al.~\cite{comminal_numerical_2018-1} use CFD simulations to model the flow of extruded material in additive manufacturing.
They study the extrusion and deposition of highly viscous material with different settings of parameters such as nozzle velocity or extrusion velocity.
This demonstrates the feasibility of using CFD simulations to model the flow behavior of TIM materials. \\
CFD simulations require a definition of the material behavior, which may be complex.
Thermal paste is typically made up of two components, e.g. a silicone grease filled with aluminum oxide particles.
In such a case, the viscosity may change both with shear stress and filler ratio~\cite{prasher_thermal_2006}.
The rheology of TIM has further been studied (see e.g.~\cite{lin_rheological_2009, sinh_thermal_2012}). 
CFD simulations typically aim to model both complex material behavior and geometries accurately. \\
Gu et al.~\cite{gu_novo_2018} create training data from a Finite Element (FE) model.
They modify the distribution of two materials within a composite material structure and solve for mechanical properties such as toughness and strength.
They train both a linear model and a Convolutional Neural Network (CNN) on this data and speed up computation times by a factor of 250, while maintaining sufficient accuracy.
Koeppe et al.~\cite{koeppe_efficient_2018} also train an ANN on data from an FE model.
They calculate the mechanical stress for a lattice structure at given load conditions.
The computation time for a single FE simulation is approximately 5-10 hours, while the ANN takes less than one second. 
They use 85 training examples to train an ANN, which has 16 output features. \\
A major limitation of all proposed approaches is the efficient generation of a training dataset that is sufficient for the design of complex ANNs.
Those prerequisites are difficult to fulfill with experiments or state-of-the-art numeric simulation models.
Therefore, we use our proposed heuristic flow behavior model to create a sufficiently high number of training samples.
\section{Simulation methods}\label{sec:simulation_methods}
\begin{figure}[!t]
	\centering
	\includegraphics[width=1\columnwidth]{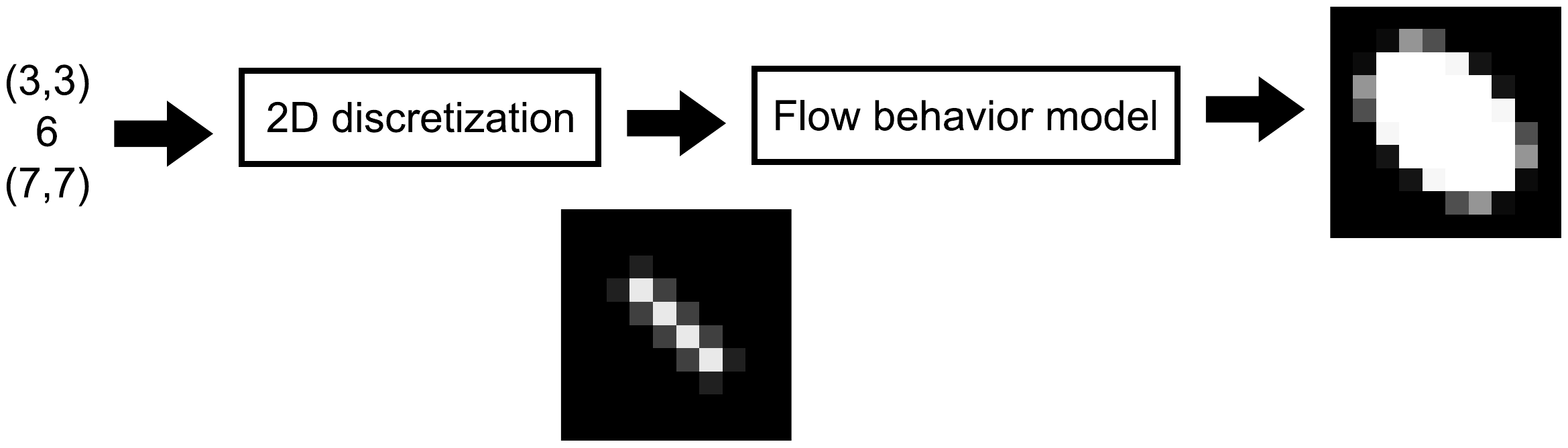}
	\caption{Overview of our approach for a single line of TIM. Inputs are the start point coordinate, feed rate and end point coordinate. The TIM distribution is spatially discretized before and after the compression step.}
	\label{fig:approach}
\end{figure}

In this section, we give an overview of our approach as depicted in Fig.~\ref{fig:approach}.
First, we describe the in- and output data representation and the 2D discretization, which is used as a pre-processing step.
We then take a closer look at each of our proposed flow behavior models: a heuristic and an ANN trained on data from the heuristic.
Specific setup details regarding e.g. dataset generation and training procedure are outlined in Section~\ref{sec:experimental_setup}. \\ \\

\subsection{Data representation and 2D discretization}
\begin{figure}[!t]
	\centering
	\includegraphics[width=.5\columnwidth]{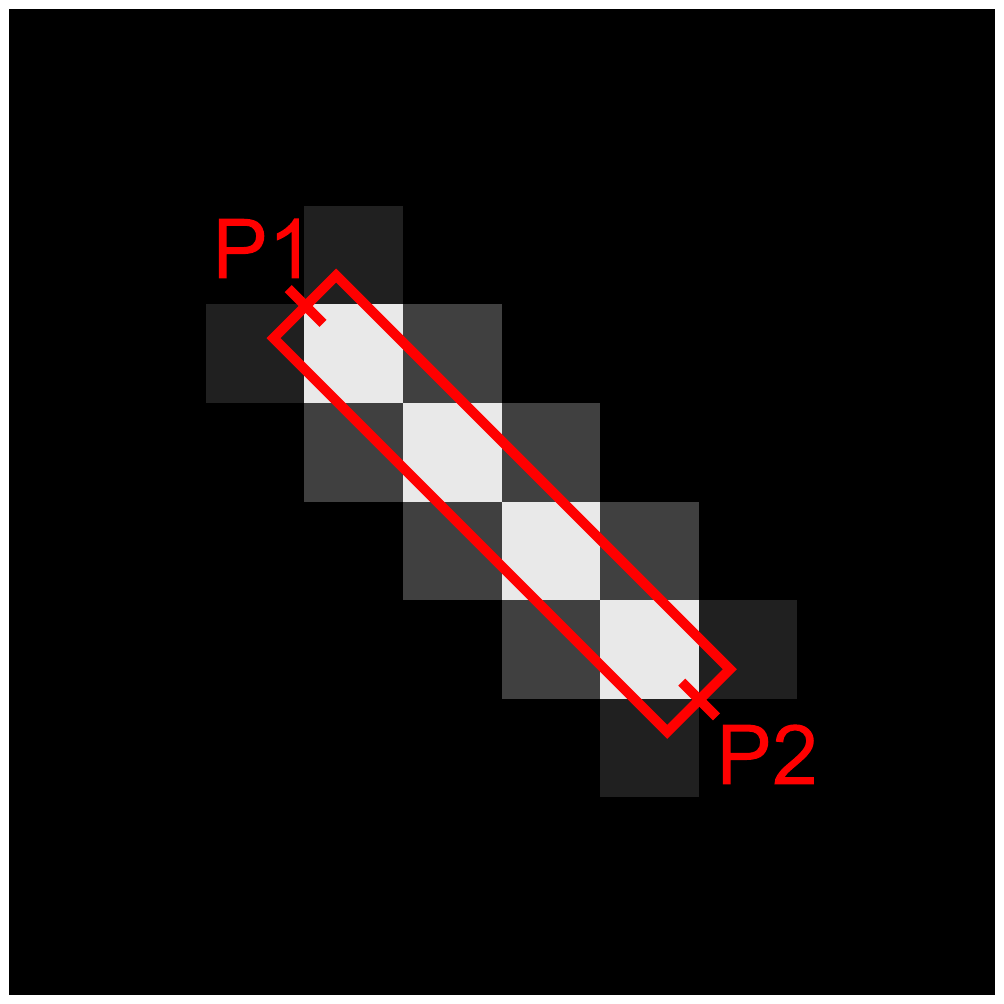}
	\caption{Visual representation of the 2D discretization for a line segment}
	\label{fig:visual_discretization}
\end{figure}

We define the dispense pattern, which is the path along which TIM is applied, using a polygonal chain.
In the simplest case, this equals a single line with five parameters:
both endpoints of this line have continuous x- and y-coordinates.
The fifth parameter is the feed rate of TIM along the line segment.
For longer patterns, we iteratively append another point and a respective feed rate.
The input parameters are shown on the left-hand side of Fig.~\ref{fig:approach}.
The state after dispensing is represented as a two-dimensional grid.
The number on each grid cell represents the amount of TIM in each cell.
The previously introduced input parameterization is transferred onto this two-dimensional representation.
This process is visualized as \textit{2D discretization} in Fig.~\ref{fig:approach}.
We apply \textit{Unweighted Area Sampling}~\cite{hughes_computer_2014}, which is a technique in the field of computer graphics to draw anti-aliased lines.
It works as follows in our case:
each segment of the pattern is assigned a width of one, i.e. all lines become rectangles.
The intersection of each grid cell with each rectangle is calculated.
The amount, which is specified via the feed rate for each line segment, is assigned to each grid cell proportional to this intersection.
Fig.~\ref{fig:visual_discretization} contains a visual depiction of how the amount for each grid cell is calculated.
This discretized state of the spatial TIM distribution after dispensing serves as input to a flow behavior model, which outputs the state after compression.
This output is again a spatial distribution of TIM and is discretized in the same way.
An example is shown on the right-hand side of Fig.~\ref{fig:approach}.
The flow behavior model can be either the heuristic or an ANN.
Both take the dispensed state as input and output the compressed state.
\subsection{Heuristic}
\begin{algorithm}[!t]
	\caption{Pseudocode of our heuristic}\label{alg:pseudocode}
		\begin{algorithmic}[1]
			\STATE {\textsc{COMPRESS}}(initial)
			\STATE \hspace{0.5cm}artificial\_height = maximum(initial)
			\STATE \hspace{0.5cm}compressed = initial
			\STATE \hspace{0.5cm}\textbf{while}(artificial\_height \textgreater~1)
			\STATE \hspace{0.5cm}~reduce\_artificial\_height()
			\STATE \hspace{0.5cm}~\textbf{while}(max(compressed) \textgreater~artificial\_height)
			\STATE \hspace{0.5cm}~~temp = zeros(max(x\_coords), max(y\_coords))
			\STATE \hspace{0.5cm}~~\textbf{for} x \textbf{in} x\_coords
			\STATE \hspace{0.5cm}~~~\textbf{for} y \textbf{in} y\_coords
			\STATE \hspace{0.5cm}~~~~diff = compressed[x,y] - artificial\_height
			\STATE \hspace{0.5cm}~~~~\textbf{if}(diff \textgreater~0)
			\STATE \hspace{0.5cm}~~~~~compressed[x,y] -= diff
			\STATE \hspace{0.5cm}~~~~~temp[next\_neighbors(x,y)] += diff $/$ 4
			\STATE \hspace{0.5cm}~~compressed += temp
			\STATE \hspace{0.5cm}\textbf{return} compressed
		\end{algorithmic}
\end{algorithm}

\begin{figure}[!t]
\centering
\includegraphics[width=1\columnwidth]{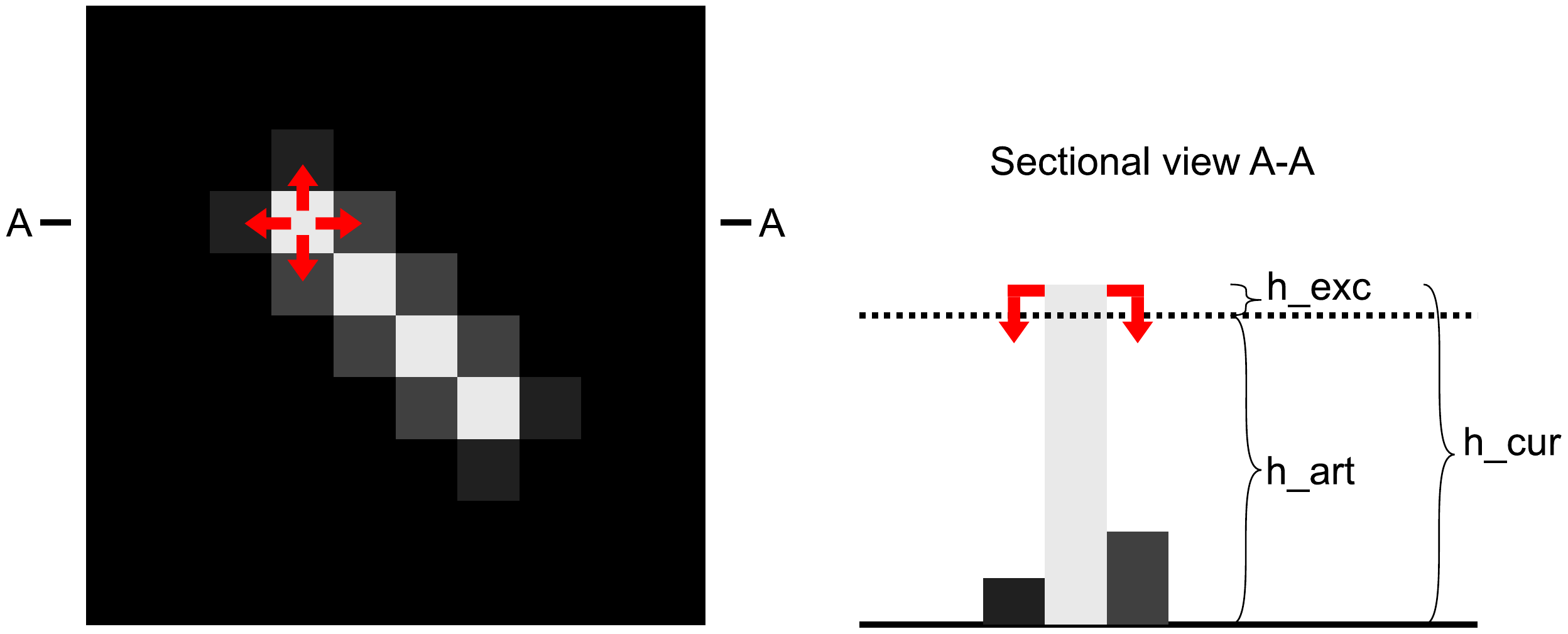}
\caption{Visual representation of an exemplary iteration of the heuristic. Left: top view; right: sectional view A-A.}
\label{fig:visual_heuristic}
\end{figure}

We now look into the details of our proposed heuristic.
Algorithm~\ref{alg:pseudocode} contains the respective pseudocode.
Fig.~\ref{fig:visual_heuristic} visualizes how the material spreads to neighboring cells during a single iteration of our algorithm. 
First, we define an artificial height value~$h_{art}$ corresponding to the maximum TIM amount.
We then enter a loop that is executed until we reach a final~$h_{art}$ equal to one.
During this loop, we iteratively reduce~$h_{art}$.
While the TIM amount in any grid cell exceeds the current~$h_{art}$, we loop over all grid cells.
For each cell, we check its current TIM amount~$h_{cur}$ against~$h_{art}$.
If~$h_{cur} > h_{art}$, we divide the excess amount $h_{exc} = h_{cur} - h_{art}$ by four and add it to each of the next neighboring cells in a temporary array.
We subtract $h_{exc}$ from the current cell of the compressed state.
After we have looped over every cell, we update the compressed state with the temporary array.
This avoids that the order of the cells within the loop has an influence on the result.
Increasing the dispensed amount on the input side has the same effect as compressing down to a lower gap height.\\
\subsection{Artificial Neural Network}
The ANN is trained on data generated by the heuristic flow behavior model.
As such, an advantage with regard to accuracy can not be expected.
However, it can map the complex relationship between in- and output more efficiently.
Programming libraries such as \textit{Keras}~\cite{chollet_keras_2015} conveniently implement ANNs ready to be executed in parallel on a Graphical Processing Unit (GPU).
The computation can be executed quickly.
Furthermore, gradient information is provided. \\
ANNs can be made up of various types of layers.
Well-known architectures such as \textit{VGG}~\cite{simonyan_very_2014}, \textit{ResNet}~\cite{he_deep_2015} or \textit{Inception}~\cite{szegedy_rethinking_2016} rely on the use of convolutional layers followed by dense layers.
They have proven to work very well with image data.
Since our data can be interpreted as gray scale images, we opt to work with a similar architecture.
Details regarding the architecture definition and the training process are presented in Section~\ref{sec:experimental_setup}. \\
\section{Experimental setup}\label{sec:experimental_setup}
This section contains information on how we set up our models and experiments.
This includes the performance benchmarking for both flow behavior models.
For the ANN, we describe the generation of the training data, the architecture design and the training process.
For the experimental data, we describe the laboratory setup.
\subsection{Training the ANN}
\begin{figure}[!t]
	\centering
	\includegraphics[width=.8\columnwidth]{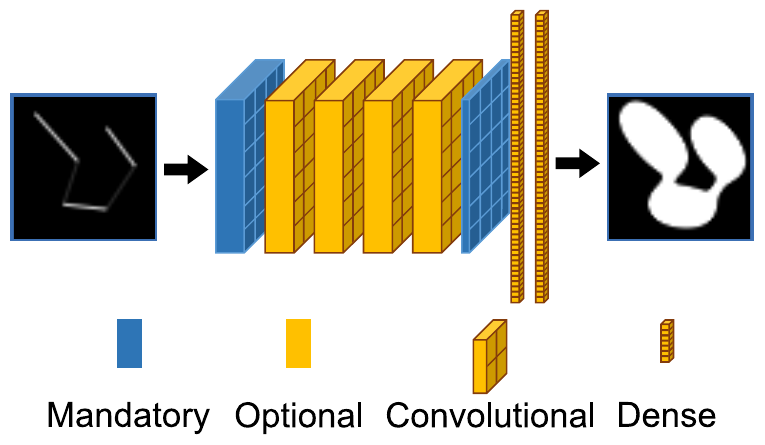}
	\caption{Architecture of the Artificial Neural Network (ANN). Hyperparameters such as the number of optional layers (yellow) are optimized. Mandatory layers (blue) are always included.}
	\label{fig:ANNarchitecture}
\end{figure}

The training dataset consists of 200\,000 automatically generated random dispense patterns.
The patterns are similar to the ones used during benchmarking as depicted in Fig.~\ref{fig:randompaths}.
We obtain the architecture of the ANN from an automated hyperparameter optimization.
A template for the architecture is visualized in Fig.~\ref{fig:ANNarchitecture}.
The layers indicated in blue are always used.
Yellow layers are optionally activated by the optimizer.
We vary the number of convolutional layers from two to six and the number of dense layers from zero to two.
The convolutional layers have either 8, 32, 128 or 512 filters with a kernel size of either three or five.
If present, each dense layer has 2500 neurons.
The batch size may be 8, 32 or 128.
The optimizer to train the ANN is \textit{Adam}~\cite{kingma_adam:_2014} with a learning rate between $10^{-5}$ and $10^{-2}$.
We use the activation function \textit{ReLu} for all layers except for the last, where we apply the \textit{Sigmoid} function.
The loss function to be optimized is \textit{binary cross-entropy}.
The weights of the ANN are initialized randomly.
Therefore, training an ANN multiple times on the same dataset yields fluctuating results.
Preliminary manual trials of architecture tuning have shown convergence issues during the training of some hyperparameter configurations.
To account for fluctuating performance and convergence issues, we train 10 ANNs for each configuration during the hyperparameter optimization.
We return the lowest loss over the 10 respective runs value back to the optimizer.
We use the hyperparameter optimization framework \textit{Optuna}~\cite{akiba_optuna:_2019}.
The hyperparameter optimization runs for 1\,000 iterations.
To make a high number of iterations possible, we train within each iteration for one epoch on 16\,000 patterns and validate on 4\,000 patterns.
After finishing the hyperparameter optimization, we fine-tune the ANN by training on 160\,000 patterns for 10 epochs and validate using 40\,000 patterns. \\
When using a different resolution, the ANN needs to be retrained as just described. \\
The ANN is trained for a constant output gap height.
However, a change in the gap height has the same effect on the result as changing the input amounts.
When using the ANN, different gap heights can thus be accounted for by scaling the input amounts respectively. \\
Since the ANN has a fixed input format, the ANN input dimensionality is set according to the maximum pattern length.
To process shorter patterns, we simply append pattern segments with zero amounts up to the maximum length. \\

\subsection{Physical experiments}\label{sec:physical_experiments}
\begin{figure}[!t]
	\centering
	\includegraphics[width=2.5in]{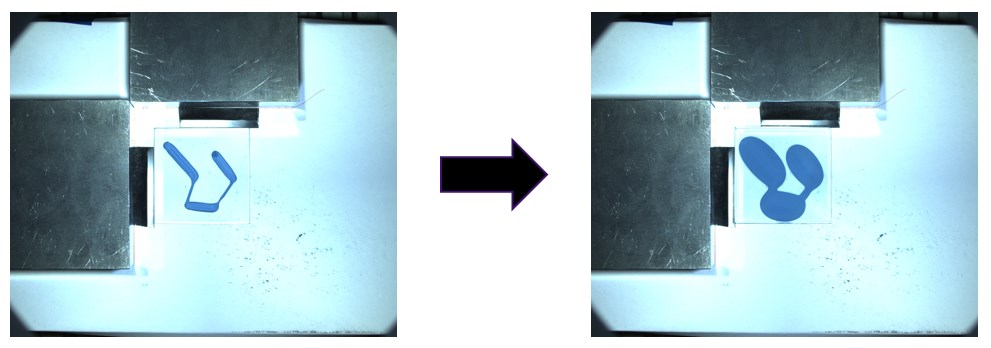}
	\caption{Laboratory experiments using transparent glass plates to compress TIM. Left: state before compression; right: state after compression.}
	\label{fig:laboratoryexperiment}
\end{figure}

To validate our model, we carry out physical experiments.
We dispense TIM in various different patterns and compress it as when joining a heatsink. \\
The machine used for dispensing is an automated Computerized Numerical Control (CNC) machine.
Its type is almost identical to the ones that are used in automotive series production.
We transfer the patterns into G-code, which is a format that is readable on this kind of machine.
The TIM is dispensed onto glass plates with a dimension of 70$\times$70\,mm.
Thin metal plates with a carefully machined height are put on the edges of the glass plate.
This ensures a uniform final gap height when putting a second glass plate on top and pushing it downwards.
The dispensed and compressed states of TIM are shown for an exemplary pattern in Fig.~\ref{fig:laboratoryexperiment}.
An image of the compressed state is recorded.
An automatic segmentation of the blue color hue is applied and yields a representation in the same discretized format as introduced previously.
Thus, each pixel is either entirely full or empty.
Since the final gap height is low, the error at the area boundary made by this assumption is sufficiently small.
After segmentation, the resolution is scaled down to the same resolution as in the heuristic model.
During downscaling, we apply a linear interpolation between neighboring cells.
The zoom level is adjusted uniformly for all experiments.
This is done as a post-processing step and has the same effect as adjusting the vertical camera position.
Since the experiments are carried out manually, some samples are shifted slightly.
Those translational errors are corrected by re-centering each pattern during post-processing. 
The post-processing does, of course, not involve a modification of the overall pattern shape, since this would distort the error evaluation of the model. \\
We further evaluate the TIM flow behavior in a physical experiment using a real product sample.
This product involves a Printed Circuit Board (PCB) with mounted electronic components to be cooled.
The housing is pressed onto the PCB and TIM spreads over both joining partners to form a thermal connection.
In contrast to our laboratory experiments as just described, we cannot control the actual gap height that precisely in this case:
multiple parts are joined together, with each part having individual mechanical tolerances.
Furthermore, the PCB itself bends during the joining process.
We thus use this experiment for a qualitative rather than a quantitative assessment.
For comparison with our model we choose the total TIM amount to fit the actually observed amount.
Instead of evaluating the general fit of our model, we evaluate the fit with respect to only the shape of the predicted coverage area outline.

\subsection{Benchmarking}
\begin{figure}[!t]
	\centering
	\includegraphics[width=.9\columnwidth]{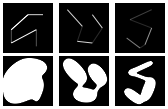}
	\caption{Three exemplary patterns used in our computation time and error benchmarking. Top row: before compression, bottom row: after compression.}
	\label{fig:randompaths}
\end{figure}

During this study, we use a resolution of 50$\times$50 cells. 
This resolution was selected through preliminary trials with different resolutions and represents a compromise between sufficient accuracy and computational effort. \\
One part of our benchmarking covers the error of our simulation models.
This involves the comparison of the entire simulation pipeline consisting of the discretization and either flow behavior model to the physical experiments.
It further involves the error between the outputs of the heuristic and the ANN.
In all cases, we calculate the absolute error of the respective compressed states:
\begin{equation}
e_{comp} = \sum_{i=1}^{50} \sum_{j=1}^{50} | m_{a,comp,ij} - m_{b,comp,ij} |,
\end{equation}
with $m_{a,comp,ij}$ and $m_{b,comp,ij}$ being the TIM amounts per grid cell $(i,j)$ in the compressed states.
The indices $a$ and $b$ refer to either the experiment and a flow behavior model or the heuristic and the ANN.
We then divide the absolute error by the sum of the covered cells
\begin{equation}\label{eq:erelsingle}
e_{rel} = \frac{e_{comp}} {\sum_{i=1}^{50} \sum_{j=1}^{50} m_{a,comp,ij}}
\end{equation}
and calculate its mean
\begin{equation}\label{eq:erelmean}
\overline{e}_{rel} = \frac{1}{N_{pat}} \sum_{k=1}^{N_{pat}} e_{rel,k}
\end{equation}
across $N_{pat} = 50$ dispense patterns.
This relative error measure yields a better intuition of the model accuracy across the different dispense patterns. \\
Besides the error, we also evaluate the computation speed for our model.
Both flow behavior models are called from a Python script.
The library \textit{timeit}~\cite{python_software_foundation_timeit_2022} measures the computation time of a code snippet.
Setup code, such as code for loading data and models, is executed separately and not included into the measurement.
Background processes may interfere with the program being measured and spuriously lengthen the computation time.
For this reason, it is specifically not recommended to report mean and average computation times for multiple runs of the same code~\cite{python_software_foundation_timeit_2022}.
Thus, we execute $N_{runs} = 10$ runs per measurement and store the minimum value 
\begin{equation}
t_{min} = \min t_{l}, l \in \{1, ... , N_{runs}\}
\end{equation}
for further evaluation.
Since measurement time varies for different patterns, we measure the computation time for the compression of $N_{pat}$ individual patterns.
Examples are shown in Fig.~\ref{fig:randompaths}.

We report the mean value
\begin{equation}
\overline{t} = \frac{1}{N_{pat}} \sum_{n=1}^{N_{pat}} {t}_{min,n}
\end{equation}
for the computation time $t_{min,n}$ across $N_{pat} = 50$ paths and the respective standard deviation
\begin{equation}
s = \sqrt{\frac{1}{N_{pat}-1} \sum_{n=1}^{N_{pat}} ({t}_{min,n} - \overline{t})^{2} }.
\end{equation}
The computation time $t_{min,n}$ for an individual pattern is, as just described, the minimum time across 10 runs per individual pattern.
All computations are executed on a workstation with an INTEL E5-2680 processor and four GPUs of type NVIDIA RTX 2080Ti. \\
\section{Results}\label{sec:results}
\begin{figure}[!t]
	\centering
	\includegraphics[width=2.5in]{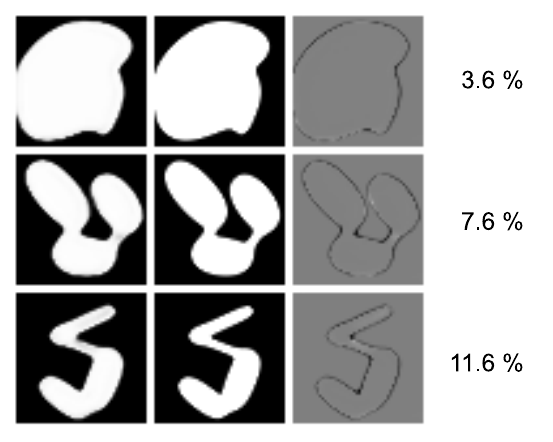}
	\caption{Output of the ANN as compared to data from the heuristic model. Left column: output of ANN for different dispense patterns, center column: output from heuristic, right column: difference of both and error score after Equation~\ref{eq:erelsingle}.}
	\label{fig:ANNheuristic}
\end{figure}

This section contains the results regarding the heuristic, the ANN and the physical experiments. 
We record the setup and computation time for all three approaches and calculate the relative absolute error of the compressed states as described previously. 
Results are listed in Table~\ref{tab:model_comparison}.\\
First, we give a deeper insight into the process of setting up the ANN.
We determine the hyperparameters by carrying out a hyperparameter optimization as described in Section~\ref{sec:experimental_setup}.
We obtain the following architecture:
the first layers are five convolutional layers with 128 filters and a filter size of 5$\times$5.
They are followed by the mandatory convolutional layer with one filter and a filter size of 3$\times$3.
No dense layers are appended.
The best remaining hyperparameters are a batch size of 8 and a learning rate of 0.0011.
The entire hyperparameter optimization process with 1\,000 iterations takes one week.
The fine-tuning of the final architecture takes about 2 hours.
It takes about one week to create the training dataset for the ANN, which involves the simulation of 200\,000 patterns. 
Those steps need to be carried out only once for a specific input resolution. \\
We compare the trained ANN with the original heuristic approach.
The error according to Equation~\ref{eq:erelmean} is \textbf{5\,\%}.
Fig.~\ref{fig:ANNheuristic} shows the output of the trained ANN as compared to the heuristic.
While some errors are prevalent in the outermost cells, the ANN manages to fit the data well. \\
We now look closer into the laboratory experiments, which are carried out as described in Section~\ref{sec:physical_experiments} and form an independent test dataset with unseen patterns.
An exemplary sample of those dispense patterns is shown in Fig.~\ref{fig:randompaths}.
Each experiment takes 30\,minutes.
This time includes sample preparation, dispensing, compression and post-processing of the results.
The experiments serve as ground truth and therefore are listed with an error equal to zero.
We are aware that they are still subject to error sources such as mechanical tolerances or measurement noise. \\
For both of our flow behavior models, we calculate the mean relative error with respect to the experiments according to Equation~\ref{eq:erelmean}.
The heuristic and the ANN are both able to predict the compressed shape well, with an error of \textbf{11\,\%} and \textbf{13\,\%} respectively across the 50~evaluated patterns.
A visual comparison of the ANN with the experiments is presented in Fig.~\ref{fig:experimental_validation}.
Further samples are shown in the appendix.
The left column shows the compressed state as output from the ANN for three different dispense patterns.
The middle column shows the compressed state acquired from the experiments.
The right column shows the difference of both along with the error score after Equation~\ref{eq:erelsingle}.
Errors occur mainly in outermost cells of the covered area. \\
Patterns with high errors are often characterized by the entrapment of air.
An example is shown in Fig.~\ref{fig:void}, where the relative error according to Equation~\ref{eq:erelsingle} is 21.4\,\%.\\
The initial dispense pattern can be calculated rather straightforward from the pattern parameters as described previously.
The setup time for simulating a certain dispense pattern is therefore relatively low and takes up to one minute.
This procedure is equal for both flow behavior models. \\
The computation time amounts to \textbf{3.41\,s} on average for the heuristic and displays a rather large variance across different patterns.
The ANN can be executed consistently in \textbf{0.07\,s}.
\begin{table}[!t]
	\centering
	\caption{Comparison of simulation methods during deployment in manual pattern design}\label{tab:model_comparison}
	\begin{tabular}{K{22mm} X{17mm} X{14mm} X{17mm}}
		\textbf{Method} & \textbf{Mean relative error} & \textbf{Setup time} & \textbf{Computation time} \\
		 & $\overline{e}_{rel}$ &   &  $\overline{t}$ \\
		\rule{0pt}{11pt} \textit{Experiment} & 0 & 30\,$min$ & - \\
		\rule{0pt}{11pt} \textit{CFD} & - & 10\,-\,60\,\,$min$  & 60\,-\,120\,$min$ \\
		\rule{0pt}{11pt} \textit{Numeric heuristic} & 11\,\% & 1 \,$min$ & 3.41\,$\pm$\,2.71$s$ \\
		\rule{0pt}{11pt} \textit{Neural network} & 13\,\% & 1 \,$min$ & 0.07\,$\pm$\,0.001$s$ \\
	\end{tabular}
\end{table}

\begin{figure}[!t]
	\centering
	\includegraphics[width=2.5in]{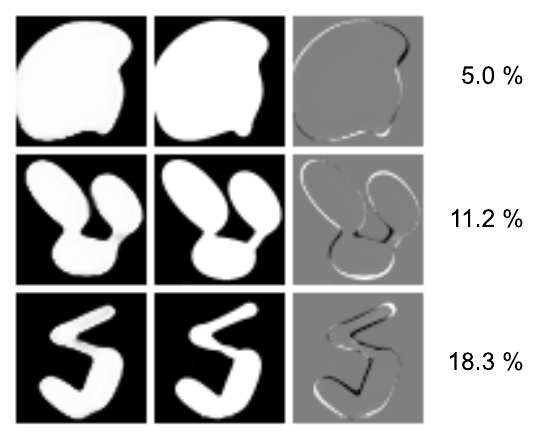}
	\caption{Validation on experimental data. Left column: output of ANN for different dispense patterns, center column: experimental data, right column: difference of both and error score after Equation~\ref{eq:erelsingle}.}
	\label{fig:experimental_validation}
\end{figure}

\begin{figure}[!t]
\centering
\includegraphics[width=2.5in]{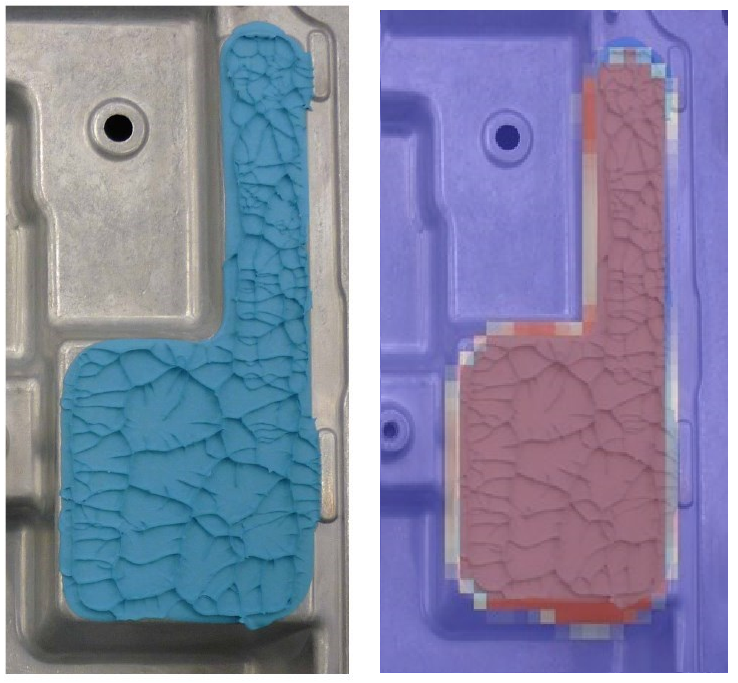}
\caption{Overlay with the compressed TIM on a real ECU. Left: image of compressed TIM in a physical experiment; right: additional overlay of the compressed state as obtained from the heuristic.}
\label{fig:ECU}
\end{figure}

\section{Discussion}\label{sec:discussion}
\begin{figure}[!t]
	\centering
	\includegraphics[width=2.5in]{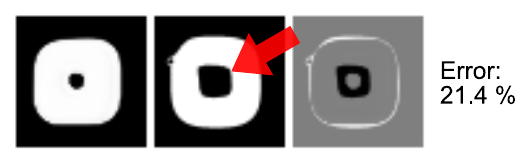}
	\caption{Dispense pattern involving a void area due to air entrapment. Left: output of ANN, center: experimental data with marked air entrapment, right: difference of both and error score after Equation~\ref{eq:erelsingle}.}
	\label{fig:void}
\end{figure}

The CFD simulations can be considered state-of-the-art in this field of simulation.
We do not aim to analyze them deeper within this work, since they have been used extensively for over 20 years in a wide range of different applications.
We have shown our samples to two experienced simulation experts and asked them for their professional opinion.
They are regularly occupied with simulations of similar dispense patterns.
They estimate the setup time for such dispense patterns to be in the range of 10\,min up to 60\,min.
10\,min would include a very basic setup without much detail.
60\,min would include a more elaborate setup, e.g. a fine modeling of 3D roundings of the dispense pattern.
The computation time is estimated by consulting the simulation logging files for similar patterns and is in the range of 60\,-\,120\,min.
Due to the high effort of CFD simulations, we omit the exact error calculation on our 50 samples. 
We do not claim that our new simulation approach offers an advantage over the CFD simulation with regard to the error. 
However regarding computational speed, they are clearly outperformed by both of our proposed surrogate flow behavior models. \\
The heuristic model captures the flow behavior of TIM during compression generally well.
It is shown that the overall shape is almost identical in most cases.
The difference to experimental trials occurs mainly at the outer areas.
The accuracy is high enough to support manual development work.
The computation time is low. \\
While the heuristic model can be executed within a few seconds, the ANN delivers results almost instantly.
The computation time improves by a factor of almost 50, but the accuracy suffers only marginally.
The application of our model can thus save a significant effort during dispense pattern design for electronics packages.
CFD simulations and physical experiments will still be necessary, but only for fine-tuning during the last design cycles.
This is specifically the case in tests involving the design limits, e.g. experiments, which cover the highest expected mechanical tolerances.
We do not claim to replace CFD simulations or experiments fully, but rather to reduce the number of trials. \\
The speed-up, which is achieved by using the ANN, allows the use not only for manual pattern design, but also for an automated pattern design.
The ANN further supplies gradient information.
Thus, the advantage regarding computational speed would be even higher, if the flow behavior model is combined with a gradient-based optimizer.
Automated pattern design with state-of-the-art optimizers could explore a much larger range of the design space and thus lead to better solutions than those found via manual trial-and-error iterations.
Several solution candidates can be generated by executing the optimization with slightly varied settings.
The design engineer can then choose the most promising patterns. \\
During training of the ANN, we only used open source software libraries.
This allows the integration of our model into a custom user interface, i.e. independently from proprietary simulation software. 
Especially a web-based implementation could provide easy access to design engineers without any license costs as seen with e.g. CFD supported tools. \\
The hyperparameter optimization supported the training process of the ANN.
Compared to preliminary manual hyperparameter tuning, the automated hyperparameter optimization yielded better results.
This is valid not only with regard to the model performance, but also with regard to the convergence behavior of the training process.
The shown hyperparameters parameterize an ANN, which can fit our dataset very well.
We thus recommend using an automated hyperparameter optimization when training on data of this kind.\\
The duration to set up the ANN is quite long.
The creation of the training dataset and the hyperparameter optimization takes almost two weeks in total.
Both processes run fully automatic and do not need any human intervention.
The advantage with regard to computation time is realized during actual usage of the model for dispense pattern design.
In contrast to the initial setup of the ANN, the duration of the design process is relevant for the time-to-market and needs to be carried out for each individual product.
It is thus beneficial to invest more time into training data generation and ANN training and in turn to speed-up the individual design process. \\
While the accuracy of our model is generally good, there is one specific drawback we would like to discuss further.
As mentioned before, we cannot predict the compressed shape well if air entrapments are present.
They are not desired in practical application, since the air significantly impairs thermal performance~\cite{gowda_voids_2004}.
Once the TIM pattern forms such an enclosed void area and is then compressed further, the entrapped air is put under pressure as well.
This pressure counteracts the material flow into the void area.
This effect is not taken into account by our model.
It can be seen in Fig.~\ref{fig:void} that our model predicts a quite small void, while the experimental data suggests that the TIM flows rather towards the outer areas than towards the center.
While this is an evident limitation of our model, it is not relevant for practical application:
since a pattern design involving voids will be neglected, an exact prediction of the compressed state is not necessary in those cases.
The predicted shape is generally still reasonable even in those cases. \\
The example of a real ECU depicted in Fig.~\ref{fig:ECU} shows that the model fits the TIM behavior not only in a laboratory environment, but also in the real product.
Our model assumes a infinitely wide planar surface and thus might model the compression behavior even beyond the physical area boundaries.
In real products, the cooling surface area is bounded and excess TIM flowing beyond the boundaries will not be compressed any further.
The experiments we conducted in the laboratory ensure a complete compression of the entire shape.
While the real product is certainly an important benchmark, the laboratory experiments give a deeper insight into the model accuracy. 
The effect of compressed material extending beyond the cooling area can be observed for example at the bottom part of Fig.~\ref{fig:ECU}.
This indicates that the dispense pattern, which is the result of a conventional pattern design process, could be improved. 
When the dispense pattern would fit the cooling area perfectly, no overflowing material would be visible. \\
\section{Conclusion}\label{sec:conclusion}
We present two flow behavior models, which can quickly predict the flow behavior of TIM when joining the heatsink.
Our proposed heuristic aids design engineers during the definition of the initial dispense pattern by providing a quick and easy method to estimate the compressed state.
This reduces the need for elaborate CFD simulations and manual experiments with product samples.
The time-to-market can thus be shortened for a variety of ECUs and power electronics components.
Training an ANN on data from our heuristic reduces accuracy only slightly, but yields a significant speed-up of computation time.
Using an ANN thus makes the manual design process even more convenient.
It further allows the efficient usage of optimizers for an automated dispense pattern optimization.
We show that the predicted compressed state fits experimental results well.
This is true not only in the laboratory, but also for a real ECU.
Future work includes the development of a method for automated dispense pattern optimization on the basis of this model.
\appendix
\section{}
\begin{figure}[!t]
	\centering
	\includegraphics[width=.95\columnwidth]{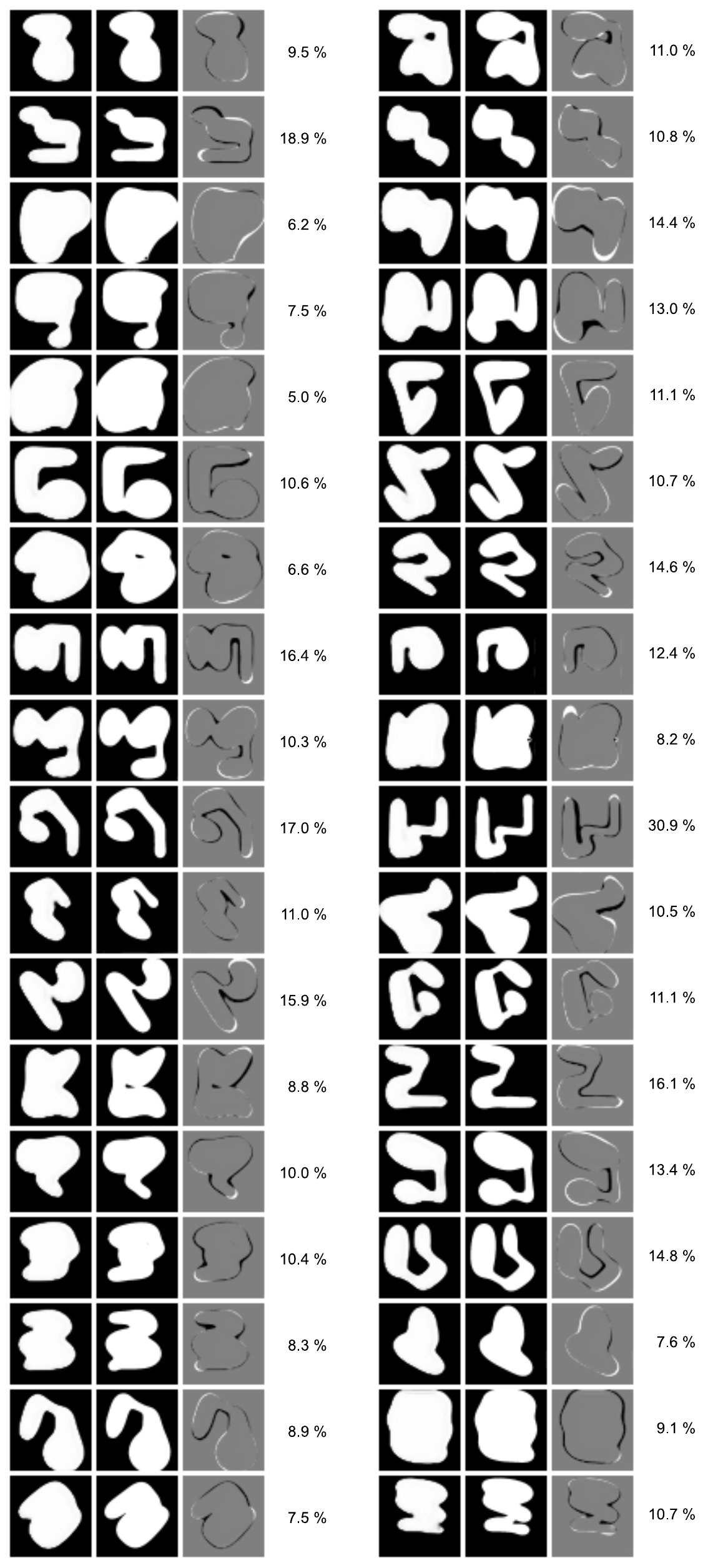}
	\caption{Validation on experimental data.  Left column: output of ANN for different dispense patterns, center column: experimental data, right column: difference of both and error score after Equation~\ref{eq:erelsingle}.}
	\label{fig:experimental_validation_fullA}
\end{figure}

In Fig.~\ref{fig:experimental_validation_fullA}, we present further examples from our experimental dataset.

%
%
%
\section*{Acknowledgment}
We thank Ralph Nyilas and Andras Horvath for the helpful discussions regarding the flow behavior of TIM.
We thank Balazs Solymossy and Istvan Horvath from the simulation department for their professional opinion regarding CFD models.
We further thank Hack-Min Kim, Vivien Reuscher and Roderich Zeiser for their support in carrying out the laboratory experiments.
\section*{Author statement}
We describe the individual contributions of Simon Baeuerle (SB), Marius Gebhardt (MG), Jonas Barth (JB), Andreas Steimer (AS) and Ralf Mikut (RM) using CRediT~\cite{brand_beyond_2015}: \textit{Writing - Original Draft}: SB; \textit{Writing - Review \& Editing}: JB, AS, RM; \textit{Conceptualization}: SB, JB, AS, RM; \textit{Investigation}: SB, MG; \textit{Methodology}: SB, AS; \textit{Software}: SB, MG; \textit{Supervision}: JB, AS, RM; \textit{Project Administration}: JB, RM; \textit{Funding Acquisition}: JB, RM.
%
%
%
%
%
\bibliographystyle{IEEEtran}
\bibliography{literature}

\begin{thebibliography}{10}
\providecommand{\url}[1]{#1}
\csname url@samestyle\endcsname
\providecommand{\newblock}{\relax}
\providecommand{\bibinfo}[2]{#2}
\providecommand{\BIBentrySTDinterwordspacing}{\spaceskip=0pt\relax}
\providecommand{\BIBentryALTinterwordstretchfactor}{4}
\providecommand{\BIBentryALTinterwordspacing}{\spaceskip=\fontdimen2\font plus
\BIBentryALTinterwordstretchfactor\fontdimen3\font minus
  \fontdimen4\font\relax}
\providecommand{\BIBforeignlanguage}[2]{{%
\expandafter\ifx\csname l@#1\endcsname\relax
\typeout{** WARNING: IEEEtran.bst: No hyphenation pattern has been}%
\typeout{** loaded for the language `#1'. Using the pattern for}%
\typeout{** the default language instead.}%
\else
\language=\csname l@#1\endcsname
\fi
#2}}
\providecommand{\BIBdecl}{\relax}
\BIBdecl

\bibitem{licari_adhesives_2011}
J.~J. Licari and D.~W. Swanson, \emph{Adhesives technology for electronic
  applications: {Materials}, processing, reliability}, 2nd~ed.\hskip 1em plus
  0.5em minus 0.4em\relax Oxford, UK: William Andrew Publishing, 2011.

\bibitem{baeuerle_cad--real:_2021}
S.~Baeuerle, M.~B{\"o}hland, J.~Barth, M.~Reischl, A.~Steimer, and R.~Mikut,
  ``{CAD}-to-real: {Enabling} deep neural networks for {3D} pose estimation of
  electronic control units,'' \emph{at - Automatisierungstechnik}, vol.~69,
  no.~11, 2021.

\bibitem{zhou_machine_2022}
B.~Zhou, T.~Pychynski, M.~Reischl, E.~Kharlamov, and R.~Mikut, ``Machine
  learning with domain knowledge for predictive quality monitoring in
  resistance spot welding,'' \emph{Journal of Intelligent Manufacturing},
  vol.~33, no.~4, pp. 1139--1163, Apr. 2022.

\bibitem{chiabert_digital_2018}
L.~F. C.~S. Dur{\~a}o, S.~Haag, R.~Anderl, K.~Sch{\"u}tzer, and E.~Zancul,
  ``Digital {Twin} {Requirements} in the {Context} of {Industry} 4.0,'' in
  \emph{Product {Lifecycle} {Management} to {Support} {Industry} 4.0}, ser.
  {IFIP} {Advances} in {Information} and {Communication} {Technology},
  P.~Chiabert, A.~Bouras, F.~No{\"e}l, and J.~Rios, Eds.\hskip 1em plus 0.5em
  minus 0.4em\relax Cham: Springer International Publishing, 2018, vol. 540,
  pp. 204--214.

\bibitem{ekpu_effects_2012}
M.~Ekpu, R.~Bhatti, N.~Ekere, S.~Mallik, and K.~Otiaba, ``Effects of {Thermal
  Interface Material}s (solders) on thermal performance of a microelectronic
  package,'' in \emph{2012 {Symposium} on {Design}, {Test}, {Integration} and
  {Packaging} of {MEMS}/{MOEMS}}, Cannes, France, Apr. 2012, pp. 154--159.

\bibitem{kesarkar_how_2019}
T.~M. Kesarkar and N.~K. Sardana, ``How {TIM} impacts thermal performance of
  electronics: a thermal point of view study to understand impact of {Thermal}
  {Interface} {Material} ({TIM}),'' in \emph{2019 {International} {Conference}
  on {Electronics} {Packaging} ({ICEP})}.\hskip 1em plus 0.5em minus
  0.4em\relax Niigata, Japan: IEEE, Apr. 2019, pp. 200--206.

\bibitem{gowda_voids_2004}
A.~Gowda, D.~Esler, S.~Tonapi, K.~Nagarkar, and K.~Srihari, ``Voids in {Thermal
  Interface Material} layers and their effect on thermal performance,'' in
  \emph{Proceedings of 6th {Electronics} {Packaging} {Technology} {Conference}
  ({EPTC} 2004)}, Singapore, Dec. 2004, pp. 41--46.

\bibitem{lee_investigation_2000}
T.-Y. Lee, ``An investigation of thermal enhancement on flip chip plastic {BGA}
  packages using {CFD} tool,'' \emph{IEEE Transactions on Components and
  Packaging Technologies}, vol.~23, no.~3, pp. 481--489, Sep. 2000.

\bibitem{comminal_numerical_2018-1}
R.~Comminal, M.~P. Serdeczny, D.~B. Pedersen, and J.~Spangenberg, ``Numerical
  modeling of the strand deposition flow in extrusion-based {Additive
  Manufacturing},'' \emph{Additive Manufacturing}, vol.~20, pp. 68--76, Mar.
  2018.

\bibitem{prasher_thermal_2006}
R.~Prasher, ``Thermal {Interface} {Materials}: {Historical} perspective,
  status, and future directions,'' \emph{Proceedings of the IEEE}, vol.~94,
  no.~8, pp. 1571--1586, Aug. 2006.

\bibitem{lin_rheological_2009}
C.~Lin and D.~D.~L. Chung, ``Rheological behavior of thermal interface
  pastes,'' \emph{Journal of Electronic Materials}, vol.~38, no.~10, pp.
  2069--2084, Oct. 2009.

\bibitem{sinh_thermal_2012}
L.~H. Sinh, J.-M. Hong, B.~T. Son, N.~N. Trung, and J.-Y. Bae, ``Thermal,
  dielectric, and rheological properties of aluminum nitride/liquid crystalline
  copoly(ester amide) composite for the application of {Thermal Interface
  Material}s,'' \emph{Polymer Composites}, vol.~33, no.~12, pp. 2140--2146,
  Dec. 2012.

\bibitem{gu_novo_2018}
G.~X. Gu, C.-T. Chen, and M.~J. Buehler, ``De novo composite design based on
  {Machine Learning} algorithm,'' \emph{Extreme Mechanics Letters}, vol.~18,
  pp. 19--28, Jan. 2018.

\bibitem{koeppe_efficient_2018}
A.~Koeppe, C.~A. Hernandez~Padilla, M.~Voshage, J.~H. Schleifenbaum, and
  B.~Markert, ``\BIBforeignlanguage{en}{Efficient numerical modeling of
  {3D}-printed lattice-cell structures using neural networks},''
  \emph{\BIBforeignlanguage{en}{Manufacturing Letters}}, vol.~15, pp. 147--150,
  Jan. 2018.

\bibitem{hughes_computer_2014}
J.~F. Hughes, \emph{Computer graphics: principles and practice}, 3rd~ed.\hskip
  1em plus 0.5em minus 0.4em\relax Upper Saddle River, NJ, USA: Addison-Wesley,
  2014.

\bibitem{chollet_keras_2015}
\BIBentryALTinterwordspacing
F.~Chollet \emph{et~al.}, ``Keras,'' 2015. [Online]. Available:
  \url{https://keras.io}
\BIBentrySTDinterwordspacing

\bibitem{simonyan_very_2014}
\BIBentryALTinterwordspacing
K.~Simonyan and A.~Zisserman, ``Very deep convolutional networks for
  large-scale image recognition,'' arXiv:1409.1556, Tech. Rep., 2014. [Online].
  Available: \url{https://arxiv.org/abs/1409.1556}
\BIBentrySTDinterwordspacing

\bibitem{he_deep_2015}
\BIBentryALTinterwordspacing
K.~He, X.~Zhang, S.~Ren, and J.~Sun, ``Deep residual learning for image
  recognition,'' arXiv:1512.03385, Tech. Rep., 2015. [Online]. Available:
  \url{https://arxiv.org/abs/1512.03385}
\BIBentrySTDinterwordspacing

\bibitem{szegedy_rethinking_2016}
C.~Szegedy, V.~Vanhoucke, S.~Ioffe, J.~Shlens, and Z.~Wojna, ``Rethinking the
  {Inception} architecture for computer vision,'' in \emph{Proceedings of the
  2016 {IEEE} {Conference} on {Computer} {Vision} and {Pattern} {Recognition}
  ({CVPR})}.\hskip 1em plus 0.5em minus 0.4em\relax Las Vegas, NV, USA: IEEE,
  Jun. 2016, pp. 2818--2826.

\bibitem{kingma_adam:_2014}
\BIBentryALTinterwordspacing
D.~P. Kingma and J.~Ba, ``Adam: {A} method for stochastic optimization,''
  arXiv:1412.6980, Tech. Rep., 2014. [Online]. Available:
  \url{https://arxiv.org/abs/1412.6980}
\BIBentrySTDinterwordspacing

\bibitem{akiba_optuna:_2019}
\BIBentryALTinterwordspacing
T.~Akiba, S.~Sano, T.~Yanase, T.~Ohta, and M.~Koyma, ``Optuna: a
  next-generation hyperparameter optimization framework,'' arXiv:1907.10902,
  Tech. Rep., 2019. [Online]. Available: \url{https://arxiv.org/abs/1907.10902}
\BIBentrySTDinterwordspacing

\bibitem{python_software_foundation_timeit_2022}
\BIBentryALTinterwordspacing
{Python Software Foundation}, ``timeit - {Measure} execution time of small code
  snippets,'' Jan. 2022. [Online]. Available:
  \url{https://docs.python.org/3/library/timeit.html}
\BIBentrySTDinterwordspacing

\bibitem{brand_beyond_2015}
\BIBentryALTinterwordspacing
A.~Brand, L.~Allen, M.~Altman, M.~Hlava, and J.~Scott,
  ``\BIBforeignlanguage{en}{Beyond authorship: attribution, contribution,
  collaboration, and credit},'' \emph{\BIBforeignlanguage{en}{Learned
  Publishing}}, vol.~28, no.~2, pp. 151--155, Apr. 2015. [Online]. Available:
  \url{http://doi.wiley.com/10.1087/20150211}
\BIBentrySTDinterwordspacing

\end{thebibliography}
\ifCLASSOPTIONcaptionsoff
  \newpage
\fi

%

%

%




\end{document}